\def\year{2021}\relax
\documentclass[letterpaper]{article} 
\usepackage{aaai21}  
\usepackage{times}  
\usepackage{helvet} 
\usepackage{courier}  
\usepackage[hyphens]{url}  
\usepackage{graphicx} 
\usepackage{natbib}  
\usepackage{caption} 
\usepackage{url}            
\usepackage{booktabs}       
\usepackage{amsfonts}       
\usepackage{nicefrac}       
\usepackage{microtype}      

\usepackage{epsfig}
\usepackage{amsmath}
\usepackage{amssymb}

\usepackage{algorithm}
\usepackage{algorithmic}

\usepackage{multirow}
\usepackage{subcaption}
\usepackage[switch]{lineno}

\newtheorem{theorem}{Theorem}

\usepackage{color}
\newcommand{\YE}[1]{{\textcolor{blue}{[#1]}}}

\title{Achieving Adversarial Robustness via Sparsity}
\author{
    Shufan Wang\thanks{Equal contribution.},
    Ningyi Liao\textsuperscript{\rm $ \ast $},
    Liyao Xiang\thanks{Joint first corresponding author.},
    Nanyang Ye\textsuperscript{\rm $ \dagger $},
    Quanshi Zhang
    \\
}
\affiliations{

    Shanghai Jiao Tong University\\

    \{wsf2146, liaoningyi, xiangliyao08, qs1022\}@sjtu.edu.cn, yn272@alumni.cam.ac.uk

}
\begin{document}

\maketitle
\begin{abstract}
    Network pruning has been known to produce compact models without much accuracy degradation. However, how the pruning process affects a network's robustness and the working mechanism behind remain unresolved. In this work, we theoretically prove that the sparsity of network weights is closely associated with model robustness. Through experiments on a variety of adversarial pruning methods, we find that weights sparsity will not hurt but improve robustness, where both weights inheritance from the lottery ticket and adversarial training improve model robustness in network pruning. Based on these findings, we propose a novel adversarial training method called {\bf inverse weights inheritance}, which imposes sparse weights distribution on a large network by inheriting weights from a small network, thereby improving the robustness of the large network.
\end{abstract}

\section{Introduction}

It is widely recognized that deep neural networks (DNNs) are usually over-parameterized, and network pruning has been adopted to remove insignificant weights from a large neural network without hurting the accuracy. Despite its success, pruning strategies have been rarely discussed in the adversarial learning setting where the network is trained against adversarial examples, and the robustness of the network is as important as accuracy.

It is unclear what pruning methods are effective and which factors are critical for retaining model robustness. Believing that the inherited model weights may not be effective in preserving network accuracy \cite{DBLP:journals/corr/abs-1903-12561, DBLP:conf/iclr/LiuSZHD19}, \citet{DBLP:journals/corr/abs-1903-12561} propose a concurrent adversarial training and weight pruning framework to seek a compressed robust model. \citet{gui2019model} further incorporates pruning and several other techniques into a unified optimization framework to preserve high robustness while achieving a high compression ratio. However, the conventional three-stage `training--pruning--fine-tuning' pipeline has not been closely examined in the adversarial context. More crucially, it is unclear which components in the network pruning methods are critical to preserving model performance. To this end, we design a comprehensive set of experiments to answer these questions.

Despite some adversarial pruning methods that have been proposed, there is still a lack of theoretical foundation to explain the working mechanism behind those methods. In fact, there are seemingly contradictory opinions on the robustness of pruned networks: \citet{DBLP:conf/iclr/MadryMSTV18} suggests network capacity is crucial to robustness, and a wider network is more likely to obtain higher accuracy and robustness than a simple network. In contrast, \citet{guo2018sparse} theoretically proves that an appropriately higher weight sparsity implies stronger robustness on naturally trained models. Other theories such as the `Lottery Ticket Hypothesis' \cite{DBLP:conf/iclr/FrankleC19} point out that, a subnetwork extracted from a large network can always achieve comparable performance with the original one in the natural setting. However, it remains unknown if the hypothesis holds true for adversarially robust networks. We are motivated to explore how adversarial pruning affects the intrinsic characteristics of the network and its impact on model robustness.

In this study, we find that the robustness of the model improves as its weights become sparser. We show that weights sparsity not only includes the traditional $L_{0}$-sparsity, {\em i.e.}, the number of parameters retained, but also a weight distribution closer to zero, represented generally by the $L_{p}$ norm of weights. These forms of sparsity can lead to robustness improvement, which is verified theoretically and experimentally.
By extensive experiments on a variety of state-of-the-art pruning methods, models, and datasets, we also demonstrate that a pruned network inheriting weights from a large robust network has improved robustness than a network with the same structure but randomly initialized weights. Moreover, weight inheritance implicitly produces sparser weights distributions on adversarially pruned models.

Inspired by the connection between model sparsity and robustness, we propose a new adversarial training strategy called {\em Inverse Weights Inheritance}: by inheriting weights from a pruned model, a large network can achieve higher robustness than being adversarially trained from scratch. The pruned model can be the `winning ticket' of the large network, as we verify that `Lottery Ticket Hypothesis' \cite{DBLP:conf/iclr/FrankleC19} holds true in the adversarial learning context. The performance results of our proposed training strategy corroborate that sparse weights and high capacity are not contradictory, but contribute joint efforts to model robustness.

The contributions of the paper can be summarized as follows. {\em First,} we establish the theoretical connection between network robustness and sparsity. {\em Second,} through comprehensive experiments, we find that weights inheritance and adversarial training are important in adversarial pruning, and implicitly provide weights sparsity. {\em Finally,} we propose a new adversarial training strategy that achieves improved robustness.

\begin{figure}[t]
	\centering
	\includegraphics[width=0.8\columnwidth]{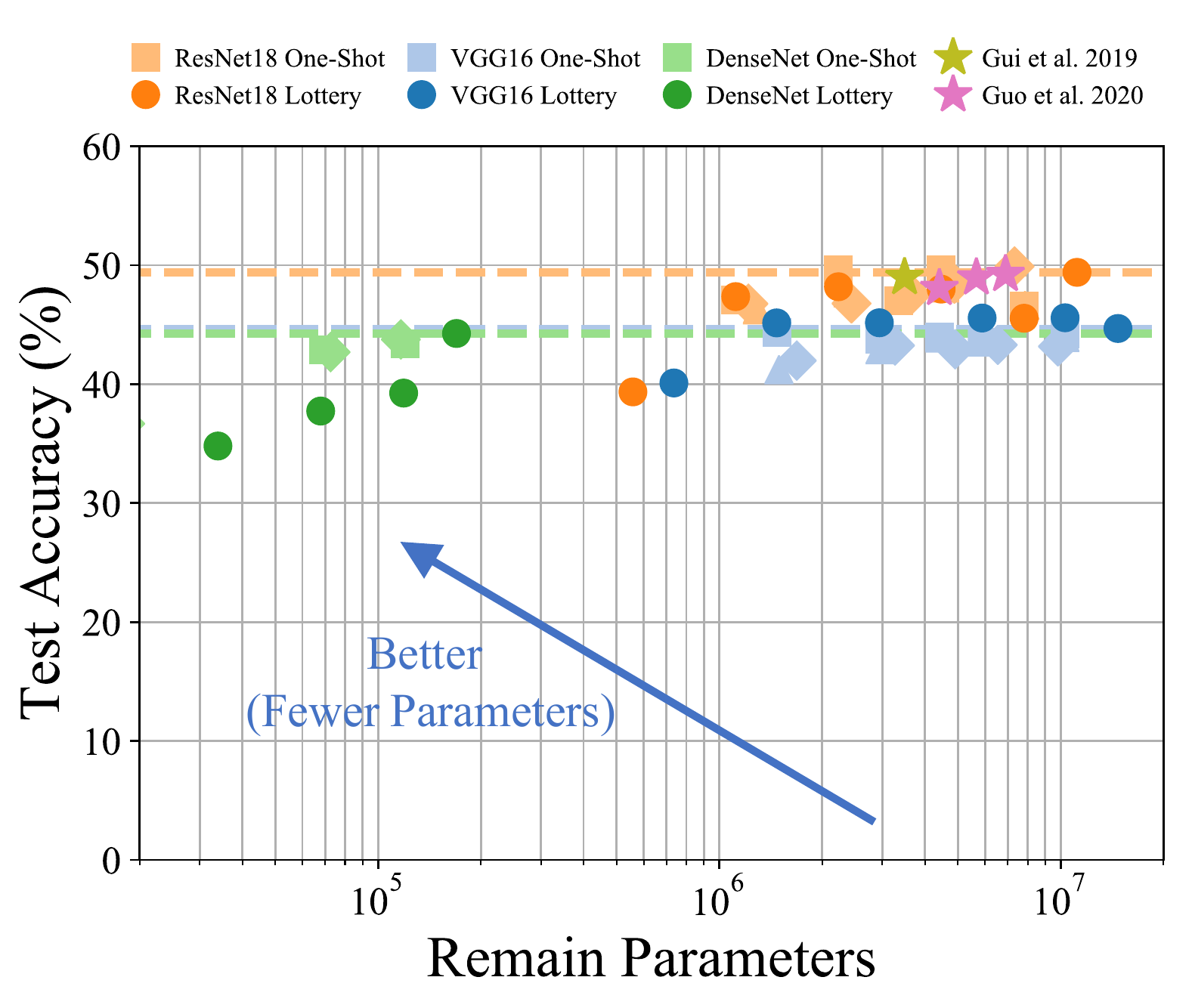}
	\caption{The relation between parameter numbers and adversarial robustness in our approach and the state-of-the-art methods on different architectures. The dotted lines represent the baselines of three base (large) models. Models residing at the upper left corner have higher adversarial accuracies and smaller sizes. All models are adversarially trained by PGD with $ \epsilon = 8/255 $ and 10 steps, and evaluated by PGD attack of $ \epsilon = 8/255 $ and 100 steps on CIFAR10. We also mark the results \citet{Guo2020} and \citet{gui2019model} by stars in the same settings. Our experiments show that adversarial pruning methods are effective in obtaining networks of smaller parameters with comparable or even better robustness than the baselines.} 
	\label{fig:nparam}
\end{figure}

\section{Related Work}


\subsubsection{Adversarial Training.} Adversarial training and its variants are proposed to improve network robustness against adversarial examples \cite{DBLP:conf/iclr/KurakinGB17}. \citet{DBLP:conf/iclr/MadryMSTV18} motivates projected gradient descent (PGD) as a universal `first-order adversary,' and optimizes the saddle point formulation to train a robust network. \citet{DBLP:journals/corr/abs-1905-09747} observes robustness can transfer between networks by knowledge distillation, and such transfer can even improve the robustness of the student network. Following the convention, we adopt $L_\infty\text{-PGD}$ attack \cite{DBLP:conf/iclr/MadryMSTV18}, {\em i.e.,} the strongest attack utilizing the local first-order information of the network, both in adversarial training strategy and the robustness evaluations.

\subsubsection{Network Pruning Methods.} Network pruning methods related to this paper can be divided into two categories: structured pruning and unstructured pruning. Structured pruning prunes a network at the level of filters \cite{DBLP:conf/ijcai/2018,DBLP:conf/iclr/0022KDSG17,DBLP:conf/iccv/LuoWL17}, channels \cite{DBLP:conf/iccv/LiuLSHYZ17} or columns \cite{wen2016learning}, depending on their respective importance. 
The importance of a filter or a channel can be determined by the norm of the weights \cite{DBLP:conf/iclr/0022KDSG17} or the channel scaling factor \cite{ye2018rethinking,DBLP:conf/iccv/LiuLSHYZ17} (sometimes the scaling factor in batch normalization layers). The unstructured pruning \cite{lecun1990optimal,hassibi1993second} prunes at the level of individual weight according to the Hessian matrix of the loss function. \citet{DBLP:journals/corr/HanPTD15} proposes to prune weights with small magnitude, and the compression ratio is further enhanced in \citet{DBLP:journals/corr/HanMD15} by quantization and Huffman coding. By incorporating non-negative stochastic gates, \citet{louizos2017learning} turns network pruning into an optimization problem with $ L_0 $-norm regularization. We pick representative structured and unstructured pruning methods to implement in our experiments. 

\subsubsection{Network Pruning in Adversarial Context.} Network pruning in adversarial context has been recently discussed in search of small and robust models \cite{wang2018adversarial,zhao2018compress,DBLP:journals/corr/abs-1903-12561,Sehwag2019}.
Several frameworks \cite{rakin2019robust,Madaan2019,gui2019model} have been proposed to adversarially train a neural network while constraining its size by pruning and/or quantization. 
However, these works do not answer which pruning factors are important for robust networks, nor which pruning methods are effective.

The `lottery ticket hypothesis' \cite{DBLP:conf/iclr/FrankleC19} shows the existence of a sparse subnetwork (or `winning ticket') in a randomly initialized network that can reach comparable performance with the large network. Nevertheless, \citet{DBLP:journals/corr/abs-1903-12561} argues against the existence of `winning ticket' in adversarial settings. On the other hand, \citet{Cosentino2019} manages to acquire adversarial winning tickets of simple models without harming model robustness. 
\citet{li2020towards} further proposed an optimized learning rate schedule to boost the searching performance of lottery tickets, while demonstrating why \citet{DBLP:journals/corr/abs-1903-12561} fails to find them. 

\citet{DBLP:conf/iclr/LiuSZHD19} claims that for network pruning in the natural setting, weights inherited by unstructured and predefined structured pruning may not be useful, as it may trap the pruned network to bad local minima. We show with experiments that weight inheritance improves the robustness in the adversarial setting, which we conjecture that this is because the inverse weights inheritance embrace larger networks during training, which can help jump out of local minima and achieve better generalization performance. \citet{hein2017formal} proposes a formal guarantee of adversarial robustness in terms of the local Lipschitz constant. By building a bridge between the local Lipschitz constant and weight sparsity, \citet{guo2018sparse} considers that an appropriately higher weight sparsity on naturally trained networks implies higher robustness. \citet{dinh2020sparsity} also finds an adversarially trained network with sparser weights distribution tends to be more robust, such as EnResNet20 \cite{wang2019resnets}. Different from compression, \citet{dhillon2018stochastic} proposes dynamic sparsity as an approach to improve robustness. By supplementing the concept of `sparsity,' we found empirical evidence of the link between robustness and sparsity, as well as training strategies to boost robustness.


%

\section{Study of Robustness and Network Pruning}
\label{section:theory}



In this section, we theoretically prove that sparser weights distribution indicates an improved level of robustness. In the theoretical deduction, we assume DNNs with ReLU activation functions, but the conclusion can be generalized to a variety of models, as we verify by experiments.

We focus on nonlinear DNNs with ReLU activation functions for classification tasks as an example to study the connection between sparsity and robustness. Let us consider a multi-layer perceptron $g\left(\cdot\right)$ trained with labeled training datasets $\{\left(x_i, y_i\right)\}$. Each layer of the network is parameterized by a weight matrix $ W_d \in \mathbb{R}^{n_{d-1}\times n_d} $ and $ w_k = W_d[:,k]$ represents the weights associated with the $k$-th class in the final layer. $ \sigma $ denotes the ReLU function. Then the prediction scores of $x_i$ for class $k$ can be denoted as
\begin{equation}
\label{equation:perceptron}
g_k\left(x_i\right) = w_k^T \sigma \left(W_{d-1}^T \sigma \left( ... \sigma \left( W_1^T x_i \right) \right) \right).
\end{equation}
Let $\hat{y}=\arg\max_{k\in\{1,...,c\}}g_k\left(x\right)$ denote the class with the highest prediction score.  Assuming the classifier is Lipschitz continuous, the local Lipschitz constant of function $g_{\hat{y}}\left(x\right)-g_k\left(x\right)$ over the neighborhood of $x$ is defined as $ L^k_{q,x} = \max_{x \in B_p\left(0,R\right)} \Vert \nabla g_{\hat{y}}(x)-\nabla g_k(x) \Vert$, where $B_p\left(x,R\right)$ denotes a ball centered at $x$ with radius $R$ under $L_p$ norm. Previous works \cite{hein2017formal,guo2018sparse} have associated robustness with the local Lipschitz constant by the following theorem:
\begin{theorem}
	\label{theorem:robustness_guarantee}
	\cite{hein2017formal,guo2018sparse} Let $\hat{y}=\arg\max_{k\in\{1,...,c\}}g_k\left(x\right)$ and $\frac{1}{p}+\frac{1}{q}=1$. For any perturbation $\delta_x\in B_p\left(0,R\right)$, $p \in \mathbb{R}^+$ and a set of Lipschitz continuous functions $\{g_k:\mathbb{R}^n\mapsto\mathbb{R}\}$, the classification decision on $x'$ will not change with
	\begin{equation}
	\label{equation:robustness_guarantee}
	\left\Vert\delta_x\right\Vert_p \le \min\left\{ \min_{k\ne\hat{y}} \frac{g_{\hat{y}}\left(x\right)-g_k\left(x\right)}{L^k_{q,x}}\right\},
	\end{equation}
	where $ L^k_{q,x} = \max_{x' \in B_p\left(0,R\right)} \Vert \nabla g_{\hat{y}}(x')-\nabla g_k(x') \Vert$.
\end{theorem}

Eqn.~\eqref{equation:robustness_guarantee} has clearly depicted the relation between robustness and the local Lipschitz constant --- a smaller $L^k_{q,x}$ represents a higher level of robustness as a larger distortion can be tolerated without changing the prediction. \citet{guo2018sparse} further gives the relation between the local Lipschitz constant and the weights. We further deduct that the relation satisfies the following theorem:
\begin{theorem}
	\label{theorem:lipschitz_sparsity}
	(The robustness and weights distribution of ReLU networks.) Letting $\frac{1}{p}+\frac{1}{q}=1$, for any $x \in \mathbb{R}^n $, $k \in \{1,...,c\}$ and $q \in \{1,2\}$, the local Lipschitz constant of function $ g_{\hat{y}}\left(x\right)-g_k\left(x\right) $ satisfies
	\begin{equation}
	\label{equation:lipschitz_sparsity}
	L^k_{q,x} \le \left\Vert w_{\hat{y}}-w_k \right\Vert_q
	\prod_{j=1}^{d-1}\left({\left\Vert W_j \right\Vert_p} \right).
	\end{equation}
\end{theorem}
Note that the local Lipschitz constant is upper bounded by the product of the $L_{p}$-norm of the weights matrices. That is to say, if $\left\Vert W_j \right\Vert_p$ is small, $L^k_{q,x}$ is constrained to be small, leading to a higher level of robustness. The proof of Thm.~\ref{theorem:lipschitz_sparsity} is omitted here due to space constraint and we refer readers to the supplementary document for the detailed proof.

We have at least two interpretations of Thm.~\ref{theorem:lipschitz_sparsity}: if we let $p=0$, Eqn.~\eqref{equation:lipschitz_sparsity} is bounded by the number of non-zero weights of the model, and hence the higher the proportion of non-zero weights, the more robust the model is. On the other hand, a smaller value of $\left\Vert W_j \right\Vert_p$ suggests the distribution of weights is closer to zero. This indicates that if a model has a weights distribution closer to zero, it may be more robust than other models with the same structure. We will respectively show how the two points are supported by the experimental results.

\section{Performance Evaluation}
\label{section:experiment}

\subsection{Implementation Details}

In this part, we describe the implementation details in examining adversarially robust network pruning. To obtain objective results, we mostly follow the experimental settings in previous works \cite{DBLP:conf/iclr/LiuSZHD19,DBLP:conf/icml/YangZXK19,DBLP:journals/corr/abs-1903-12561,DBLP:conf/icml/ZhangZ19}. Our experiments are carried out with PyTorch 1.0 on NVIDIA GeForce 2080 Ti GPUs.

\subsubsection{Datasets and Networks.} For the fairness of the results, we conduct experiments on CIFAR-10, Tiny-ImageNet, and CIFAR-100, which are representatives for small-scale datasets, large-scale datasets and datasets somewhere in between. Three state-of-the-art network architectures are chosen: VGG \cite{DBLP:journals/corr/SimonyanZ14a}, ResNet \cite{DBLP:conf/cvpr/HeZRS16}, and DenseNet \cite{DBLP:conf/cvpr/HuangLMW17} as the base large networks. A DenseNet-BC with depth $40$ and growth rate $k=12$ is also used. 

\subsubsection{One-Shot Pruning Methods.} We pick four representative and intrinsically different pruning methods: Global Unstructured Pruning (\textbf{GUP}) \cite{DBLP:conf/iclr/FrankleC19}, Local Unstructured Pruning (\textbf{LUP}) \cite{DBLP:journals/corr/HanMD15}, Filter Pruning (\textbf{FP}) \cite{DBLP:conf/iclr/0022KDSG17} and Network Slimming (\textbf{NS}) \cite{DBLP:conf/iccv/LiuLSHYZ17}. LUP and GUP are unstructured pruning, whereas FP and NS are structured pruning. Both GUP and NS prune globally according to the importance of weights or channels across all convolutional layers, while LUP and FP prune an identical percentage of weights or filters per layer locally. FP is a predefined pruning method while GUP, LUP and NS are automatic pruning methods where the structure is determined by the pruning algorithm at runtime.

We conduct these pruning methods in a one-shot manner that removes the parameters at one step, followed by post retraining to convergence. For all pruning methods, we re-implement each to achieve comparable performance with that reported in the current literature. For FP in ResNet, we conduct it on every two consecutive convolutional layers and skip the shortcuts according to \cite{DBLP:conf/iccv/LuoWL17}, also it is not available on DenseNet as pruning one filter would lead to input channel changes in all subsequent layers \cite{DBLP:conf/iclr/0022KDSG17,DBLP:conf/iclr/LiuSZHD19}. For NS, the highest pruning ratio is selected according to the maximum channel pruning ratio to avoid the removal of layers \cite{DBLP:conf/iccv/LiuLSHYZ17}.

\begin{table*}[!t]
	\centering
	\caption{\textbf{Clean testing accuracy/adversarial testing accuracy (in \%)/distortion lower bound} of pruned networks with adversarial retraining. Accuracy and distortion bounds higher than the base model are in bold.}
	\fontsize{9pt}{10pt}\selectfont
	\begin{subtable}[c]{\linewidth}
		\centering
			\begin{tabular}{@{}cccccc@{}}
				\toprule
				\multicolumn{6}{c}{\textbf{(a)   } One-Shot Pruning on CIFAR-10 w/ Stop-E} \\ \toprule
				Network                                                                           & $p\%$ & LUP & GUP & FP & NS \\ \midrule
				\multirow{3}{*}{\begin{tabular}[c]{@{}c@{}}ResNet18\\ (82.84/49.40/2.519)\end{tabular}}
				& 30  & \textbf{82.13/49.9/3.221}    & \textbf{81.92}/46.56/2.402   & \textbf{83.62}/46.61/2.505   & \textbf{84.18/49.92}/2.023      \\
				& 60  & 82.21/48.44/\textbf{2.777}   & \textbf{84.73/49.64/2.612}   & 82.61/48.08/2.501   & \textbf{83.57/49.46/2.666}      \\
				& 90  & 80.09/46.76/1.533            & \textbf{83.89}/47.09/\textbf{2.940}  & 78.87/46.24/1.764   & -                \\ \midrule
				\multirow{3}{*}{\begin{tabular}[c]{@{}c@{}}VGG16\\ (78.57/44.68/3.471)\end{tabular}}
				& 30  & \textbf{79.81}/43.17/1.982  & \textbf{80.43}/44.24/1.630    & 77.05/43.91/3.002   & \textbf{80.10}/43.81/2.991       \\
				& 60  & \textbf{78.78}/43.30/2.136  & \textbf{80.26}/43.51/2.275    & 77.13/44.21/2.032   & \textbf{79.56}/44.29/2.607      \\
				& 90  & 72.1/41.98/2.510            & \textbf{79.83}/44.36/2.501    & 69.38/41.20/2.270   & \textbf{79.54}/43.76/2.443      \\ \midrule
				\multirow{3}{*}{\begin{tabular}[c]{@{}c@{}}DenseNet-BC\\ (76.01/44.26/1.109)\end{tabular}}
				& 30  & 74.42/43.76/\textbf{1.525}  & 74.68/43.40/\textbf{2.928}   & -             & 73.86/43.08/\textbf{2.572}              \\
				& 60  & 73.16/42.70/\textbf{1.734}  & 73.24/42.88/\textbf{1.781}   & -             & 66.33/37.54/1.059              \\
				& 90  & 63.15/36.68/\textbf{2.000}  & 65.19/36.85/\textbf{1.784}   & -             & -              \\ \bottomrule
			\end{tabular}
	\end{subtable}
	\begin{subtable}[c]{.9\linewidth}
		\centering
			\begin{tabular}{@{}cccccc@{}}
				\toprule
				\multicolumn{6}{c}{\textbf{(b)   } One-Shot Pruning on Tiny-ImageNet w/ Stop-C} \\ \toprule
				Network                                                                           & $p\%$ & LUP    & GUP   & FP          & NS          \\ \midrule
				\multirow{3}{*}{\begin{tabular}[c]{@{}c@{}}ResNet18\\ (41.94/14.43/2.594)\end{tabular}}
				& 30  & \textbf{42.72/14.87}/2.356 & \textbf{43.18/15.82/2.713} & \textbf{42.68/14.91/3.300} & \textbf{41.89/15.23}/2.397 \\
				& 60  & \textbf{42.28/15.51/3.022} & \textbf{42.80/16.12/3.272} & 40.88/\textbf{15.87}/1.172 & 37.92/14.11/\textbf{3.250} \\
				& 90  & 40.32/\textbf{16.11}/2.581 & \textbf{42.21/17.26/2.797} & 36.79/14.43/1.819 & -         \\ \midrule
				\multirow{3}{*}{\begin{tabular}[c]{@{}c@{}}DenseNet121\\ (49.48/19.65/1.922)\end{tabular}}
				& 30  & \textbf{48.86/20.03/3.616}   & \textbf{48.19/20.52/2.519} & -  & 46.43/\textbf{19.71}/1.597        \\
				& 60  & \textbf{48.63/19.98/2.300} & \textbf{48.96/19.92/1.984} & -           & 40.82/16.51/\textbf{3.334}         \\
				& 90  & 45.72/18.75/1.722         & 46.99/18.65/1.478 & -           & -
				\\ \bottomrule
			\end{tabular}
	\end{subtable}
	\label{table:advcom_three_datasets}
\end{table*}

\subsubsection{Adversarial Training and Evaluation.} We employ the widely used $ L_\infty\text{-PGD} $ adversary with $ \epsilon = 8/255, \text{step size} = 2/255 $ in our experiments. Following recent works \cite{Guo2020}, we utilize $ \text{iteration} = 10 $ for adversarial training, and evaluate robustness on $ \text{iteration} = 100 $. For all trainings, we adopt a SGD optimizer with momentum of $ 0.9 $ and weight decay of $ 5\times 10^{-4} $. The batch sizes for CIFAR-10 and CIFAR-100 are both $128$. On Tiny-ImageNet, the batch size is $128$ for ResNet18 and $32$ for DenseNet121 following \citet{DBLP:conf/icml/ZhangZ19} and \citet{DBLP:conf/icml/YangZXK19}. The distortion bound of adversarial examples \cite{bastani2016measuring, salman2019convex} also serves as a robustness metric, which is estimated by searching the minimum PGD $\epsilon$ that crafts a valid adversarial image on a given batch. We report the average of distortion bounds across all samples.



\subsubsection{Stopping Criteria.} Typically, it is not well-defined how to train models to `full convergence' when stepwise decaying learning rate schedule is applied. Hence we adopt two stopping criteria indicating models have been sufficiently trained for ease of comparison. \textbf{Stop-E} denotes the network is trained for a fixed number of epochs. For CIFAR-10, CIFAR-100, and Tiny-ImageNet, we set the start learning rate to be $0.1, 0.1,$, and $0.01$, respectively. The learning rate is divided by $ 10 $ for every $1/3$ of the total epochs. \textbf{Stop-C} monitors the validation loss changes to automatically adjust the learning rate. For example, if we define patience to be $ 10 $ epochs and relative threshold to be $ 10^{-5} $, the learning rate only decays when the average validation loss does not decrease by more than $ 0.001\%$ for consecutive $10$ epochs. Models stop training after $2$ learning rate decays.



\subsection{Adversarial Network Pruning Improves Robustness by Imposing Higher Sparsity}
\label{section:story_1}

\begin{figure*}[t]
	\centering
	\includegraphics[width=1.0\linewidth]{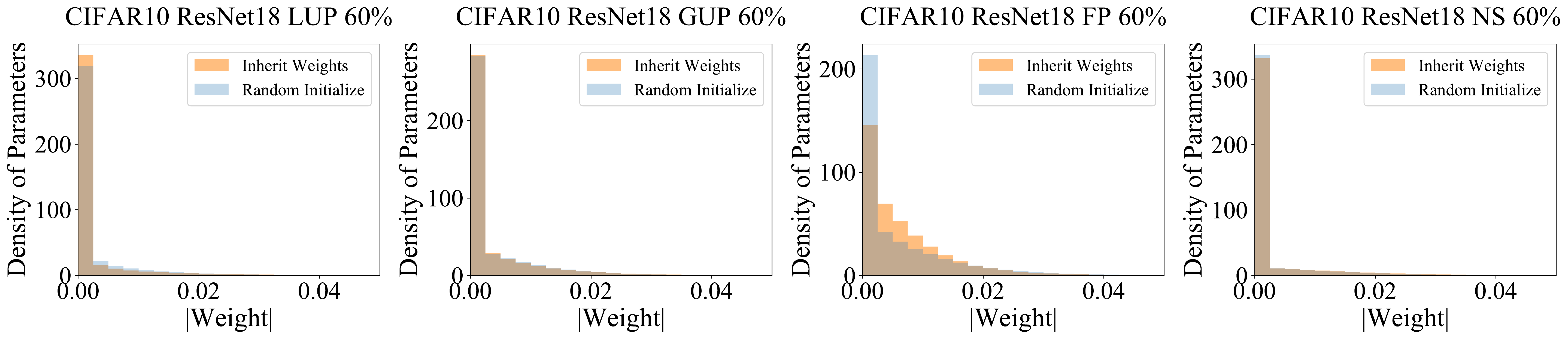}
	\caption{Weights distribution of the pruned network adversarially trained with \textbf{inherited weights} or \textbf{randomly initialized} weights. In general, networks with inherited weights from automatic pruning methods including LUP, GUP, NS have an equivalent or higher sparsity than their counterparts with randomly initialized weights. FP has lower sparsity than FP-rand.}
	\label{fig:weight_distribution_reset_inherit}
\end{figure*}

Although Thm.~\ref{theorem:lipschitz_sparsity} establishes a preliminary link between sparsity and robustness, it does not tell us how to achieve sparsity and therefore robustness by the equation. An intuitive way is to prune a network to reduce the number of non-zero weights of the model, which is also done in \cite{guo2018sparse} but only in the natural setting. We show in the following that pruning also works in the adversarial setting. Beyond that, we found that adversarial retraining after pruning mostly improves robustness, at a sparser weights distribution than models with the same structure.

We first adversarially train each base network until reaching the state-of-the-art clean and adversarial accuracy, and then prune each network by different means. 

Although pruning shows a promising method to introduce sparsity, it does not end up in robust models each time. We hence impose adversarial retraining on pruned networks to enhance robustness. The results are provided in Table~\ref{table:advcom_three_datasets}. Since there is a tradeoff between accuracy and robustness \cite{DBLP:conf/icml/ZhangYJXGJ19}, and some models tend to sacrifice one for the other, we choose to report the performance where the sum of adversarial accuracy and clean accuracy is the highest. Distortion bound is also reported for a complete view. We refer readers to the supplementary material for further discussions on the results.

\begin{table}[!t]
	\centering
	\caption{\textbf{Clean testing accuracy/adversarial testing accuracy} (in \%) of scratch networks that shares the same structure with the corresponding pruned network, only with the weights randomly initialized. Accuracy and distortion bounds higher than the base model are in bold.}
	\fontsize{9pt}{10pt}\selectfont
		\begin{tabular}{@{}c@{~}c@{~}c@{~~}c@{~~}c@{~~}c@{}}
			\toprule
			\multicolumn{6}{c}{\begin{tabular}[c]{@{}c@{}}One-Shot Pruning w/ Randomly Initialized Weights\\on CIFAR-10 w/ Stop-E\end{tabular}}                                                                                                                                                                    \\ \toprule
			Network                   & $ p\% $ & LUP-rand             & GUP-rand   & FP-rand            & NS-rand     \\ \midrule
			\multirow{3}{*}{ResNet18}
			& 30  & 81.91/46.54  & 81.41/48.78  & 82.24/46.58  & \textbf{83.21}/46.28 \\
			& 60  & 82.58/46.40  & 81.91/47.46  & 80.64/46.22  & 82.61/46.04 \\
			& 90  & 78.36/44.97  & 81.99/46.90  & 80.34/45.50  & - \\ \midrule
			\multirow{3}{*}{VGG16}
			& 30  & \textbf{79.67}/42.37  & \textbf{79.56/45.26}  & \textbf{80.27}/44.38  & \textbf{78.68}/43.52 \\
			& 60  & 78.26/43.79  & \textbf{80.23/45.22}  & 78.52/44.21  & \textbf{78.77}/43.74 \\
			& 90  & 72.32/41.38  & 77.78/43.97  & 74.24/42.39  & 77.37/43.52 \\ \bottomrule
		\end{tabular}
	\label{table:compare_prune_and_scratch}
\end{table}

Most networks in Table~\ref{table:advcom_three_datasets} obtain higher accuracy and robustness than pruning without retraining, and a large proportion of them can achieve better performance than the base networks. Specifically, LUP and NS only suffer notable performance degradation at high pruning ratios, whereas GUP remains a remarkable high performance across all pruning ratios. FP cannot preserve network performance well.

To see whether the weights inherited from a large network help the pruned network converge, we conduct a series of comparison experiments, as shown in Table~\ref{table:compare_prune_and_scratch}. Compared to FP, FP-rand initializes a small network with the same structure as the corresponding pruned network. For automatic pruning methods including LUP, GUP, and NS, we re-use the pruned network structure with re-initialized weights. As we found, compared with FP-rand, FP provides little or no improvement with the inherited weights. On the contrary, automatic pruning with inherited weights almost always performs better than that with randomly initialized weights.

Although Table~\ref{table:advcom_three_datasets}, and Table~\ref{table:compare_prune_and_scratch} experimentally found effective methods or factors to gain robustness for pruned networks, it still remains unclear how it relates to sparsity. Interestingly, by examining the weights distribution after adversarial retraining, we found that most automatically pruned networks with inherited weights have similar or higher sparsity than those with randomly initialized weights, as some examples shown in Fig.~\ref{fig:weight_distribution_reset_inherit}, while the networks pruned by predefined pruning (FP) show the opposite trend. This could be explained by Thm.~\ref{theorem:lipschitz_sparsity}, since a weight distribution closer to zero implies higher robustness. Therefore, weight inheritance and adversarial retraining implicitly provide a way to obtain sparse networks. 

\subsubsection{Comparison with previous results.}
We also compare our conclusion with previous works and summarize the difference as follows. We find inherited weights by automatic pruning (LUP, GUP, NS) provide better initialization for small networks, while predefined pruning does not. \citet{DBLP:conf/iclr/LiuSZHD19} argues that weights inherited from structured pruning have little impact on the performance of the pruned network. While the experiments on FP agree with the conclusion, that on NS does not. \citet{wang2018adversarial} also suggests inherited weights are important to preserving network accuracy and robustness in adversarial settings, but they do not discuss the working mechanism behind. 



\begin{algorithm}[!b]
	\renewcommand{\algorithmicrequire}{\textbf{Input:}}
	\renewcommand{\algorithmicensure}{\textbf{Output:}}
	\caption{Lottery Ticket in Adversarial Settings}
	\fontsize{9.5pt}{10.5pt}\selectfont
	\label{alg:lottery_ticket}
	\begin{algorithmic}[1] 
		\REQUIRE A large network $ f \left(x;\theta_0\odot M_0\right)$ where $x$ is the input, $\theta_0$ is the randomly initialized weights, and $M_0=1^{|\theta_0|}$ denoting weight masks. \\
		~~~~~ Iterative pruning ratio $ p\% $. \\
		~~~~~ Pruning iteration $ K $. \\
		~~~~~ Training epochs $ N $ per pruning iteration. \\
		\ENSURE A winning ticket $ f \left(x;\theta_0\odot M_K\right) $.
		\FOR{k in $ \left\{1,\dots,K\right\} $}
		\STATE {Conduct adversarial training on $ f \left(x;\theta_{0} \odot M_{k-1}\right)$ for $ N $ epochs and obtain the network $ f \left(x;\theta_k\odot M_{k-1}\right)$.}
		\STATE {Prune $ p\% $ weights from the current network and obtain a new weights mask $ M_k $.}
		\STATE {Re-initialize weights of $ f $ as $ f \left(x;\theta_0\odot M_k\right) $.}
		\ENDFOR
		\STATE{\textbf{return} $ f \left(x;\theta_0\odot M_K\right) $.}
	\end{algorithmic}
\end{algorithm}

\begin{table}[!t]
	\centering
	\caption{\textbf{Clean testing accuracy/adversarial testing accuracy} (in \%) of adversarially trained `winning ticket.' $p\%$ is the pruning ratio. `60 ($ 20\times \text{3 iter} $)' means iteratively remove $20\%$ of the weights in each for 3 iterations to achieve a final pruning ratio of $60\%$. Each iteration of pruning is preceded by 1 epoch of training, and the total training epoch is 240. Accuracy and distortion bounds higher than the base model are in bold. }
	\fontsize{9pt}{10pt}\selectfont
		\begin{tabular}{@{}cccc@{}}
			\toprule
			\multicolumn{4}{c}{ Winning Tickets on CIFAR-10 and CIFAR-100 w/ Stop-E }   \\ \toprule
			Network & $p\%$ & CIFAR-10            & CIFAR-100 \\ \midrule
			\multirow{6}{*}{ResNet18}
			& 0 (baseline) & 82.84/49.40 & 50.50/21.13 \\
			& 30 ($ 30\times \text{1 iter} $) & \textbf{84.29}/45.54 & \textbf{50.62/21.72} \\
			& 60 ($ 20\times \text{3 iter} $) & \textbf{84.03}/47.99 & \textbf{52.68/21.54}  \\
			& 80 ($ 20\times \text{4 iter} $) & \textbf{81.41}/48.19 & \textbf{52.23}/20.80  \\
			& 90 ($ 30\times \text{3 iter} $) & 70.29/47.36 & 49.43/21.27 \\
			& 95 ($ 31.7\times \text{3 iter} $) & 70.29/39.35 & - \\ \midrule
			\multirow{6}{*}{VGG16}
			& 0 (baseline) & 78.57/44.68 & 44.44/18.86 \\
			& 30 ($ 30\times \text{1 iter} $) &  \textbf{80.90/45.58} & 42.21/\textbf{19.16} \\
			& 60 ($ 20\times \text{3 iter} $) &  \textbf{80.05/45.56} & 42.65/\textbf{19.12}  \\
			& 80 ($ 20\times \text{4 iter} $) &  \textbf{79.30/45.16} & 45.90/\textbf{18.93}  \\
			& 90 ($ 30\times \text{3 iter} $) &  \textbf{78.87/45.15} & 45.76/\textbf{18.89} \\
			& 95 ($ 31.7\times \text{3 iter} $) & 68.48/40.10 & -  \\
			\bottomrule
		\end{tabular}
	\label{table:lottery_ticket}
\end{table}

\begin{table*}[!t]
	\centering
	\caption{\textbf{Clean testing accuracy/adversarial testing accuracy (in \%)/distortion lower bound}. Performance of base networks are marked under the model name. The performance of the inherited Winning Tickets is shown in Table~\ref{table:lottery_ticket}. Performance higher than the base model is in bold.}
	\fontsize{9pt}{10pt}\selectfont
		\begin{tabular}{@{}cccc|cccc@{}}
			\toprule
			\multicolumn{4}{c}{\textbf{(a)} Inverse Weights Inheritance on CIFAR-10}  &\multicolumn{4}{|c}{\textbf{(b)} Inverse Weights Inheritance on CIFAR-100}\\ \toprule
			Network & $p\%$ & Stop-C & Stop-E  & Network & $p\%$ & Stop-C & Stop-E         \\ \midrule
			\multirow{4}{*}{\begin{tabular}[c]{@{}c@{}}ResNet18\\ 82.84/49.40/2.519 \end{tabular}}
			& 80  & \textbf{84.05/50.30/2.728}   & \textbf{83.14/49.59}/2.303
			& \multirow{4}{*}{\begin{tabular}[c]{@{}c@{}}ResNet18\\ 50.50/21.13/3.047\end{tabular}}
			& 30  & \textbf{53.49/22.07/3.872} & \textbf{51.57}/20.68/1.191\\
			& 90  & \textbf{83.56/49.89/2.819}   & 81.68/49.03/\textbf{2.662}  & & 60  & \textbf{52.98/21.65/3.197} & \textbf{50.74/21.28/3.472}\\
			& 95  & \textbf{84.60}/49.34/\textbf{2.659}   & \textbf{83.19}/48.93/1.734  & & 80  & 50.06/21.03/1.900 & \textbf{50.78/21.15}/1.881\\
			& -   & -                            & -                           & & 90  & \textbf{52.91/21.53}/2.812 & 50.16/\textbf{21.39}/2.056\\
			\midrule
			\multirow{4}{*}{\begin{tabular}[c]{@{}c@{}}VGG16\\ 78.57/44.68/3.471 \end{tabular}}
			& 80  & \textbf{81.21/47.38}/2.338   & \textbf{81.15/47.46}/1.600
			& \multirow{4}{*}{\begin{tabular}[c]{@{}c@{}}VGG16\\ 44.44/18.86/2.338\end{tabular}}
			& 30  & \textbf{47.18/18.91}/1.491          & \textbf{44.79/19.10}/1.534\\
			& 90  & \textbf{81.36/47.54}/2.597   & \textbf{80.74/47.53}/2.622  & & 60  & \textbf{45.97/19.38}/1.147          & \textbf{45.16}/18.68/1.191\\
			& 95  & \textbf{81.29/46.98}/3.341   & \textbf{80.68/47.59}/3.219  & & 80  & \textbf{46.51/19.22/2.631}          & 43.79/\textbf{18.90/2.759}\\
			& -   & -                            & -                           & & 90  & \textbf{47.64/19.26/3.169} & 43.30/\textbf{18.94}/2.000\\
			\bottomrule
		\end{tabular}
	\label{table:iwi}
\end{table*}

\subsection{Lottery Tickets in Adversarial Settings}

We seek that in a randomly-initialized large network, if a subnetwork exists achieving comparable robustness as the large one, which is also known as the `winning ticket' in \citet{DBLP:conf/iclr/FrankleC19} in a natural setting. More specifically, we perform Alg.~\ref{alg:lottery_ticket} to find out the `winning ticket' in the adversarial setting. A discussion of hyperparameters can be found in the supplementary material.

The results on CIFAR-10 and CIFAR-100 are displayed in Table~\ref{table:lottery_ticket}. We mark the results with comparable performance to the base large networks in bold. On ResNet18 and VGG16 trained on CIFAR-10, no noticeable performance degradation occurs when the pruning ratio is as high as $ 80\% $. This is slightly different from pruning on natural models \cite{DBLP:conf/iclr/FrankleC19}, where accuracies do not drop until pruning ratio reaches around $ 88.2\% $ and $ 92 \% $ respectively on Resnet18 and VGG16. We think the difference may be explained by the more complicated decision boundary of a robust model (in theory, a model with a higher Rademacher complexity is needed to achieve adversarial robustness), and hence its `winning ticket' requires a higher capacity.

\begin{figure}[h]
	\centering
	\includegraphics[width=\columnwidth]{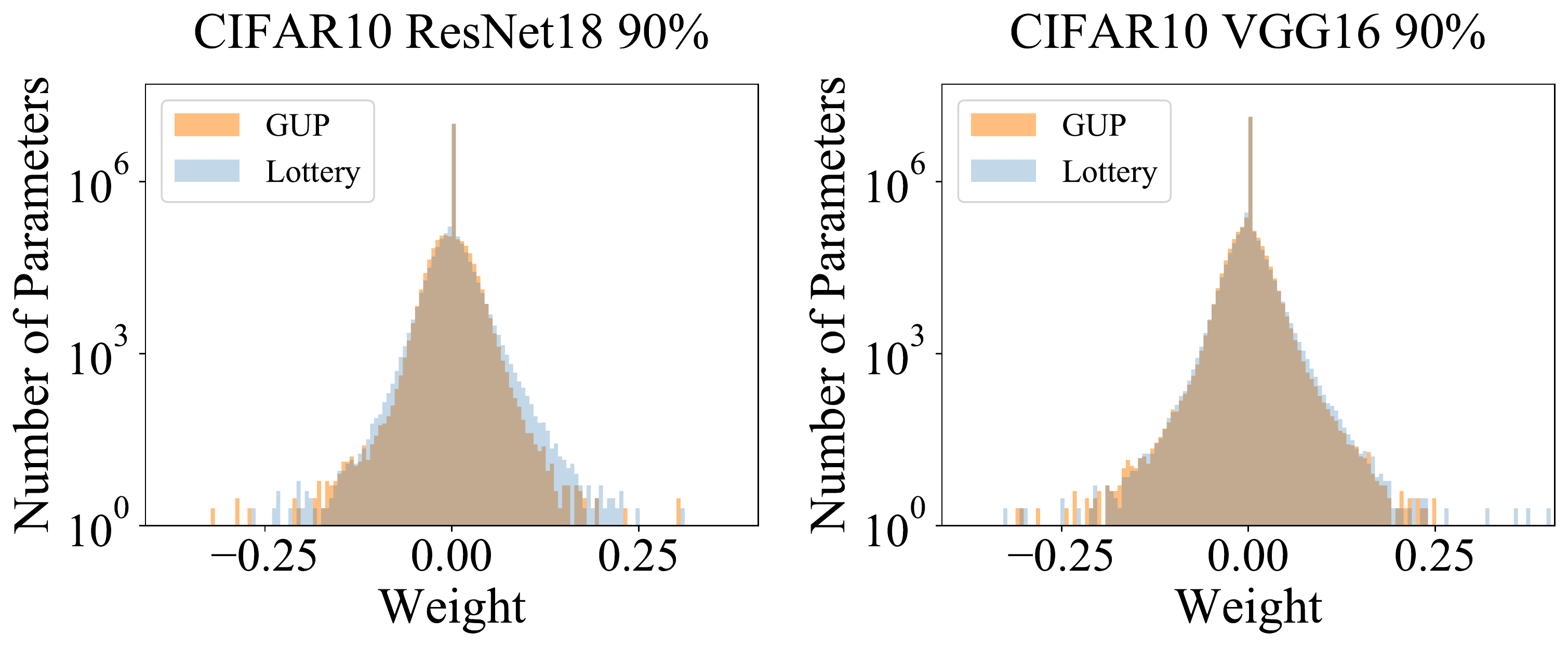}
	\caption{Weights distribution example of the pruned network obtained by one-shot \textbf{GUP} and \textbf{adversarial lottery} at the same pruning ratio. Note that we have a logarithmic y-axis such that the near-zero values are highly dense in the upper part of the figure. The distribution indicates that the adversarial winning tickets have higher sparsity than corresponding GUP pruned models. }
	\label{fig:lottery_gup}
\end{figure}

To better understand lottery tickets in adversarial settings, we compare the weights distribution between one-shot pruned model and the winning ticket at the same pruning ratio. Fig.~\ref{fig:lottery_gup} illustrates the example of two models pruned at the same pruning ratio by GUP and Alg.~\ref{alg:lottery_ticket} respectively on CIFAR10, with adversarial accuracy 47.09\% versus 47.36\% on ResNet18, and 44.36\% versus 45.15\% on VGG16, correspondingly. 
As we observe, whereas GUP models tend to have a flatter distribution which is consistent with \citet{DBLP:journals/corr/abs-1903-12561}, the winning tickets have more near-zero valued weights, indicating a higher level of sparsity. Thus we conclude that it is able to achieve preferable adversarial robustness through the lottery tickets settings. 

\subsubsection{Comparison with previous results.}
\citet{DBLP:journals/corr/abs-1903-12561} argues against the existence of `winning ticket' in adversarial settings. Nevertheless, through experiments we show that `winning ticket' exists in adversarial settings and can be obtained efficiently with a few rounds of pruning and less retraining. Our conclusion is different mostly because we search `winning ticket' by iterative global unstructured pruning as in \citet{DBLP:conf/iclr/FrankleC19}, while \citet{DBLP:journals/corr/abs-1903-12561} uses a layer-wise pruning method. As indicated in \citet{DBLP:conf/iclr/FrankleC19}, layers with fewer parameters may become bottlenecks under a layer-wise pruning method, and thus winning tickets fail to emerge. We also compare our work with \citet{li2020towards}, and find the few-shot pruning in \citet{li2020towards} does not outperform iterative pruning results in our setting. 

We also plot the results in Table \ref{table:advcom_three_datasets} and Table \ref{table:lottery_ticket} by showing the relation between the number of parameters of the pruned models against the adversarial accuracy in Fig.~\ref{fig:nparam}. By comparing with recent works including RobNet \cite{Guo2020} and ATMC \cite{gui2019model} utilizing the same training and testing metrics, which is PGD10 and PGD100, respectively, we demonstrate that our approach are able to acquire smaller networks with robustness comparable to the original dense models through adversarial network pruning, extensively effective under different current model structures among ResNet, VGG, and DenseNet.

\begin{figure*}[t]
	\centering
	\includegraphics[width=\textwidth]{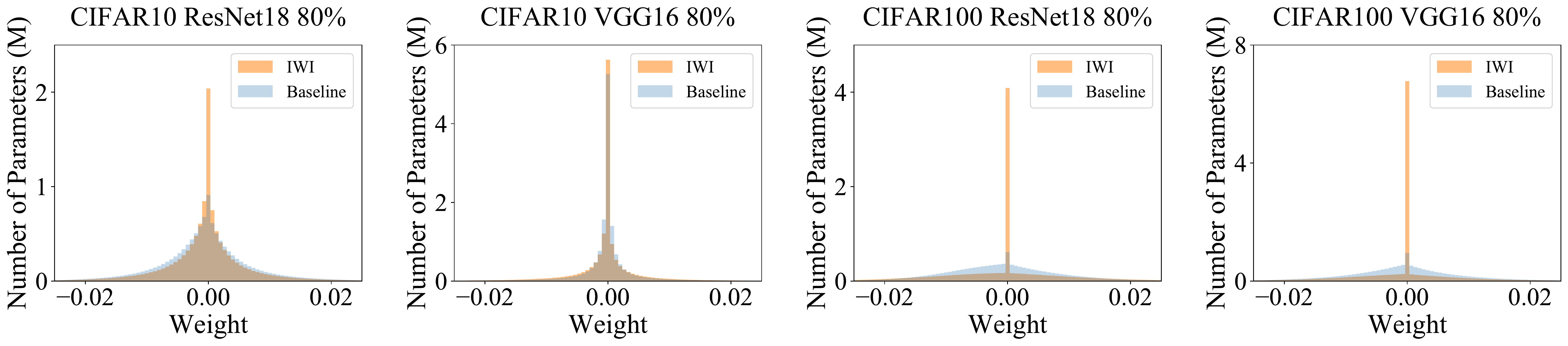}
	\caption{ Weights distribution of large networks trained by \textbf{Inverse Weights Inheritance (IWI)} and adversarially trained with \textbf{random initialization (Baseline)}. Net XX\% denotes a large network trained by inheriting the weights of a `winning ticket' with XX\% weights pruned.}
	\label{fig:weight_distribution}
\end{figure*}

\begin{algorithm}[!b]
	\renewcommand{\algorithmicrequire}{\textbf{Input:}}
	\renewcommand{\algorithmicensure}{\textbf{Output:}}
	\caption{Inverse Weights Inheritance (w/ Lottery Ticket)}
	\fontsize{9.5pt}{10.5pt}\selectfont
	\label{alg:iwi}
	\begin{algorithmic}[1] 
		\REQUIRE \{$ f \left(x;\theta_0\odot M_0\right)$, $ p\% $, $ K $, $ N $\} same as in Alg.~\ref{alg:lottery_ticket}. \\
		~~~~~ Adversarial fine-tuning epochs $ N_f $.
		\ENSURE A robust network $ f \left(x;\theta''\right) $.
		\STATE {Find the winning ticket $ f \left(x;\theta_0\odot M_K\right) $ by Alg.~\ref{alg:lottery_ticket}.}
		\STATE {Adversarially fine-tune the `winning ticket' for $ N_f $ epochs, obtain a robust small network $g \left(x;\theta'\odot M_K\right)$.}
		\STATE {Load the weights of the pruned network $g$ to the corresponding place in the large network $f$ and obtain $f \left(x;\theta'\odot M_K\right).$}
		\STATE {Re-initialize weights of $f$ as $f \left(x;\theta'\odot M_K+\theta_{0} \odot \left(M_0-M_K\right)\right).$}
		\STATE{Train $f$ until convergence and obtain $ f \left(x,\theta''\right) $.}
		\STATE{\textbf{return} $ f \left(x;\theta''\right) $.}
	\end{algorithmic}
\end{algorithm}

\subsection{Inverse Weights Inheritance}
According to our experimental results in one-shot adversarial pruning, it seems that networks with smaller capacities (higher $L_0$-sparsity) can also have an equivalent or even higher accuracy and robustness than large networks. This appears to be contradictory to the conclusion in \citet{DBLP:conf/iclr/MadryMSTV18} that classifying examples in a robust way requires the model to have a larger capacity, as the decision boundary is more complicated. We ask the question that, {\em can a network be sparse and have larger capacity at the same time?} As we analyze, it is indeed possible to have such networks with superior performance.

We introduce a new training strategy called {\em inverse weights inheritance} (\textbf{IWI}), which is inspired by Thm.~\ref{theorem:lipschitz_sparsity} and adversarial network pruning results. By the strategy, a large network acquires sparse weights distribution by inheriting weights from a small robust network, which is pruned from the same large network in the first place and is adversarially trained. Alg.~\ref{alg:iwi} gives an example of using the lottery ticket to obtain such a small network. For a fair comparison, we train the base networks with Stop-C and Stop-E (240 epochs) and report the one with higher performance. To train the large network with inherited weights, we first run Alg.~\ref{alg:lottery_ticket} to obtain the `winning ticket' and then train the `winning ticket' (a small network) for 120 epochs. Then the weights of the trained `winning ticket' are loaded back to the large network to train for another 45 epochs (Stop-E) or until convergence (Stop-C). In Table~\ref{table:iwi}, the large network with inherited weights not only outperforms the `winning ticket' but also exceeds the base network.

To find out the reason, we measure the weight distributions of each network and partial results are given in Fig.~\ref{fig:weight_distribution}. It is clear that, with inherited weights as initialization, the distribution of the final weights for the large networks is sparser (closer to zero) than those with random initialization, which is in accord with Thm.~\ref{theorem:lipschitz_sparsity}. The results suggest that for networks with the same structure, IWI implicitly finds sparse weights distribution for the large networks, and the network can achieve an improved level of clean and adversarial accuracies. Moreover, it is evident that those networks are sparse and have large capacities at the same time.

Beyond performance boost, IWI also accelerates the adversarial training process, mainly due to the lower expense of adversarially training a small network, and less training epochs required after the large network inheriting weights. Details can be found in the supplementary material. We have also tried other methods, such as using an additional regularization term to impose sparsity in large networks, but it failed. Interested readers may refer to the supplementary material for more details.

\section{Conclusion}

%

We conduct comprehensive studies on adversarial network pruning. The contributions are three-fold: First, we give a new explanation on the connection between robustness and network sparsity, which is supported by much empirical evidence. 
Second, we demonstrate the efficacy of training  network with robustness via our proposed algorithm including one-shot pruning and searching the `winning ticket.'
Third, we discover a new adversarial training strategy to achieve sparsity and large capacity at the same time for robustness.

\newpage
\bibliography{ref_cln}

\end{document}